\documentclass[10pt,twocolumn,letterpaper]{article}

\usepackage{cvpr}              
\definecolor{cvprblue}{rgb}{0.21,0.49,0.74}
\usepackage[pagebackref,breaklinks,colorlinks,allcolors=cvprblue]{hyperref}
\usepackage{multirow}
\usepackage[accsupp]{axessibility} 
\def\paperID{27} 
\def\confName{CVPR}
\def\confYear{2026}

\title{Asynchronous Remote Sensing Time-Series Fusion for Cloud Removal and Anytime Reconstruction}

\author{
Forouzan Fallah\textsuperscript{1} \quad 
Chia-Yu Hsu\textsuperscript{2} \quad
Wenwen Li\textsuperscript{2} \thanks{Corresponding author: \texttt{wenwen@asu.edu}} \quad
Anna Liljedahl \textsuperscript{3} \quad 
Yezhou Yang\textsuperscript{1} \quad  
 \\ \\ 
\textsuperscript{1} School of Computing and Augmented Intelligence, Arizona State University \\ 
\textsuperscript{2} School of Geographical Sciences and Urban Planning, Arizona State University \\
\textsuperscript{3} Woodwell Climate Research Center \\
}

\begin{document}
\maketitle
\begin{abstract}

Frequent cloud cover severely limits the usability of Sentinel-2 (S2) optical time series for Earth surface monitoring. Sentinel-1 (S1) SAR provides all-weather complementary observations, but practical S1/S2 fusion remains difficult because acquisitions are irregular and asynchronous. Many existing approaches assume temporally aligned inputs (or require external nearest-date matching) and typically restore only observed timestamps, limiting reconstruction under long gaps and preventing on-demand synthesis.
We propose AGFlow (Time-\textbf{A}ligned \textbf{G}enerative \textbf{Flow} Matching), a spatiotemporal flow-matching model for S1/S2 cloud removal and time-series reconstruction with three capabilities: (1) timestamp-conditioned internal alignment that fuses asynchronous S1 and cloudy S2 observations without preprocessing-based pairing; (2) spatiotemporal, context-aware denoising that models spatial structure jointly with temporal dynamics (rather than independent per-pixel time series); and (3) anytime querying, enabling generation of cloud-free S2 frames at both observed and user-specified timestamps within the monitoring window.
We evaluate on the RESTORE-DiT benchmark protocol with quantitative metrics, qualitative comparisons, and component ablations. AGFlow notably improves fully missing-frame reconstruction (MAE and RMSE reduce by 16--19\% over RESTORE-DiT) and provides reliable reconstructions under persistent gaps, while also yielding competitive cloud removal performance and flexible temporal querying for downstream tasks such as dense vegetation monitoring. Project page: \href{https://AGFlow-model.github.io}{AGFlow-model.github.io}
\end{abstract}
    
\section{Introduction}
\label{sec:intro}

\begin{figure*}
    \centering
    \includegraphics[width=\linewidth]{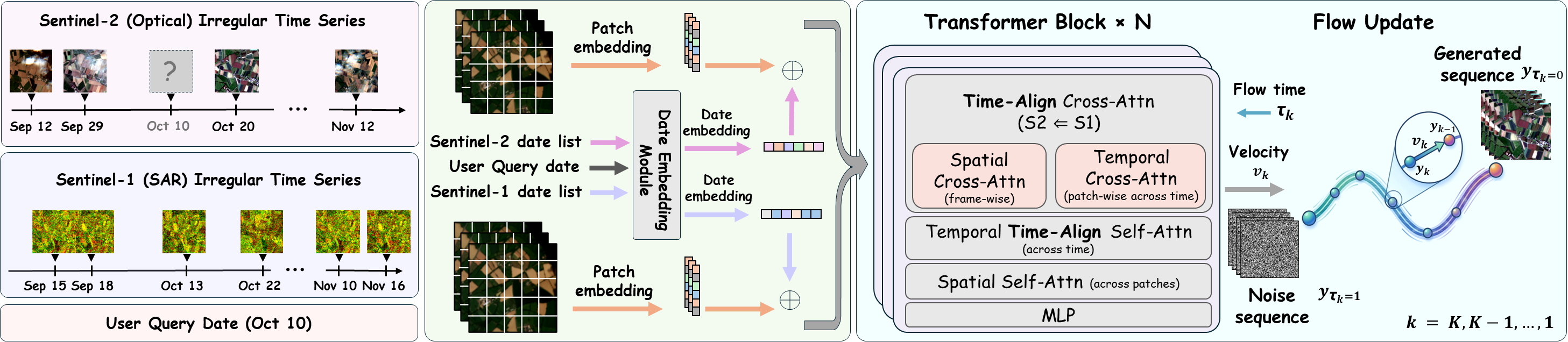}
    \caption{ Overview of AGFlow. Given an irregular Sentinel-2 optical sequence $\mathbf{x}$ with missing pixels or frames and a co-registered Sentinel-1 SAR sequence $\mathbf{s}$, images are tokenized into spatio-temporal patches and enriched with acquisition-date embeddings.
A Sequential Denoising Transformer (SDT) with time-aligned cross-attention fuses SAR features and predicts a masked velocity field $\mathbf{v}_\theta(\mathbf{z}_{\tau}, \tau, \mathrm{cond})$, where $\mathbf{z}_{\tau}=\mathbf{m}\odot\mathbf{y}_{\tau}+(1-\mathbf{m})\odot\mathbf{x}$ clamps observed pixels.
Starting from noise at ($\tau=1$) an ODE solver integrates to ($\tau=0$) while updating only masked regions, producing the reconstructed optical sequence. By changing $\mathbf{m}$, the same formulation supports both cloud/sensor-gap removal and anytime generation.
}
    \label{fig:teaser}
\end{figure*}

Dense optical satellite image time series are central for Earth monitoring. They support tracking land dynamics such as vegetation seasonality, farming activity, land-cover change, and rapid mapping after hazards. Among public missions, Sentinel-2 (S2) is widely used because it provides multi-spectral imagery at up to 10\,m resolution with global repeat coverage \cite{drusch2012sentinel}.

In practice, however, cloud and shadow reduce the usability of S2 time series, creating large spatial gaps and irregular temporal sampling. At a global scale, cloud cover is persistent and widespread (e.g., Moderate Resolution Imaging Spectroradiometer (MODIS)-derived cloud fractions on the order of $\sim$67\% globally) \cite{king2013spatial}, so even frequent-revisit sensors can experience multi-week periods with few usable observations. This breaks downstream applications that require temporally dense and spatially consistent reflectance, such as crop growth monitoring or subtle change detection.

This motivates the problem of cloud removal \cite{meraner2020cloud} and time-series reconstruction: given cloud-contaminated S2 observations and their acquisition dates, recover a cloud-free S2 series that is spatially coherent (preserving boundaries and fine structures) and temporally consistent (capturing seasonal dynamics and abrupt events). Real-world reconstruction is often challenging under persistent cloud cover, and practical sequences are irregularly sampled due to orbit geometry and quality filtering. This pushes toward models that couple spatial structure with temporal evolution rather than treating pixels independently.

A natural remedy is to integrate all-weather observations from Sentinel‑1 (S1) synthetic aperture radar (SAR), which is insensitive to clouds and illumination conditions \cite{torres2012gmes, torres2012sentinel}. However, leveraging SAR to reconstruct optical reflectance across time remains nontrivial. S1 and S2 acquisitions are typically asynchronous, and the relationship between SAR backscatter and optical reflectance is indirect, so effective fusion must handle timestamp mismatch and modality gaps without assuming a deterministic co-temporal mapping. 

Many existing SAR-optical cloud removal pipelines either assume (near-)paired S1/S2 inputs \cite{meraner2020cloud, cai2025fusing} or construct aligned sequences via external nearest-date matching \cite{shu2025restore}, which adds workflow complexity and can introduce temporal mismatches. Recent work has started to model asynchronous inputs and enable query-time decoding at user-specified timestamps, but remains pixel-wise and deterministic, limiting its use of spatial image structure \cite{tang2026anytimeformer}. Uncertainty-aware reconstruction further improves reliability, yet existing approaches are often constrained to short windows or produce only a single output frame, rather than a generative, time-conditioned model that supports dense reconstruction and arbitrary-time querying across the monitoring period \cite{cai2025fusing, liu2025effective}. 

In this work, we propose AGFlow (Time-\textbf{A}ligned \textbf{G}enerative \textbf{Flow} Matching), a timestamp-conditioned spatiotemporal flow-matching \cite{lipman2023flow} framework that addresses these practical constraints. The model treats acquisition times as first-class inputs to perform internal alignment across irregular, asynchronous S1 and cloudy S2 observations, and it supports arbitrary-time querying: generating cloud-free S2 frames at both observed S2 timestamps (cloud removal) and user-specified timestamps within the window (on-demand synthesis). By operating on spatiotemporal image representations, the denoising backbone jointly models spatial structure and temporal dynamics, while diffusion sampling provides a natural mechanism to represent uncertainty under long cloud-induced gaps.

We evaluate under the RESTORE-DiT benchmark protocol \cite{shu2025restore} and additionally assess arbitrary-time querying using a trajectory-agreement evaluation inspired by AnytimeFormer \cite{tang2026anytimeformer}.
Our main contributions are:
\begin{itemize}
    \item \textbf{Internal timestamp-conditioned alignment} for irregular and asynchronous S1/S2 fusion, eliminating preprocessing-based temporal pairing.
    \item \textbf{Spatially aware spatiotemporal conditional flow matching} for cloud removal and reconstruction, preserving spatial structure beyond independent per-pixel modeling and enabling sampling-based uncertainty.
    \item \textbf{Anytime querying} that synthesizes cloud-free S2 at both observed and user-specified timestamps within the monitoring window, unifying cloud removal, missing-frame reconstruction, and on-demand generation.
\end{itemize}

\section{Related Work}
\label{sec:related_work}

\noindent\textbf{S1/S2 cloud removal and reconstruction (paired inputs).}
A common SAR-optical formulation assumes temporally paired S1 and cloudy S2 observations, and reconstructs a cloud-free S2 image at the same timestamp. Classical CNN-based fusion methods such as DSen2-CR \cite{meraner2020cloud} adopt this paired-input assumption and output a single deterministic reconstruction from one cloudy S2 conditioned on co-temporal S1. More recent work explores conditional diffusion for single-pair restoration (e.g., EDM-CR \cite{cai2025fusing}) to improve robustness and perceptual quality in challenging scenes \cite{fallah2025rareflow}. While effective when co-temporal S1 is available, these approaches do not directly address the prevalent case where S1 and S2 are acquired asynchronously and irregularly across a monitoring window.

\noindent\textbf{Multi-temporal S1/S2 reconstruction (fixed window or aligned sequences).}
To improve robustness under persistent cloud-induced gaps, multi-temporal methods incorporate longer temporal context. Several designs operate on short or fixed-length windows (e.g., UNCRtainTS \cite{ebel2023uncrtaints}, MTS2ONet \cite{dong2024integrating}) and typically output reconstructions at observed S2 timestamps. Sequence-to-sequence models such as U-TILISE \cite{stucker2023u} target full time-series reconstruction, and can be extended with SAR inputs (U-TILISE-SAR) to use S1 as additional context. More recently, diffusion-based sequence frameworks further strengthen performance by modeling conditional generation over time (e.g., RESTORE-DiT \cite{shu2025restore}; EMRDM \cite{liu2025effective}). However, many practical pipelines still construct aligned multi-modal sequences during preprocessing (e.g., nearest-date matching of S1 acquisitions to S2 timestamps) to satisfy paired-input assumptions inside the network. Such external alignment adds workflow complexity and can introduce temporal mismatch by distorting the original acquisition timing, motivating approaches that perform alignment internally while respecting irregular and asynchronous acquisition.

\noindent\textbf{Irregular-time modeling, asynchronous fusion, and internal alignment.}
A growing line of work explicitly targets irregular sampling and/or asynchronous multi-sensor acquisition via timestamp encoding and time-aware attention. In the S1/S2 setting, AnytimeFormer \cite{tang2026anytimeformer} directly addresses irregular and asynchronous fusion and enables decoding at user-specified query times, but it models per-pixel temporal trajectories and is deterministic, which can limit spatial fidelity (e.g., boundaries and fine structures) and does not naturally represent uncertainty under long gaps. Beyond purely pixel-wise formulations, STORI \cite{liu2025innovative} incorporates local spatial neighborhood information by extracting spatiotemporal features from blocks centered on a target pixel and modeling temporal evolution; however, it still relies on local blocks rather than jointly modeling full-frame spatial structure across time, and it does not provide query-time synthesis at arbitrary timestamps. Beyond SAR-optical fusion, optical-only cross-sensor time-series works demonstrate the value of irregular-time modeling and query-time decoding in related settings (e.g., TAMRFSITS \cite{michel2026temporal}; MISR-S2 \cite{okabayashi2024cross}), but they do not address SAR-assisted cloud removal under persistent cloud gaps.

\noindent\textbf{Generative reconstruction and uncertainty.}
Deterministic reconstruction is widely used for cloud removal due to its simplicity and efficiency, but it can be brittle when observations are missing for extended periods. GAN-based approaches (e.g., MTS2ONet \cite{dong2024integrating}) can generate sharper outputs but may be harder to stabilize temporally. Diffusion models provide a strong alternative for conditional restoration and naturally support sampling-based uncertainty, which is valuable when persistent cloud cover yields underconstrained reconstructions (e.g., EDM-CR \cite{cai2025fusing}; RESTORE-DiT \cite{shu2025restore}; EMRDM \cite{liu2025effective}). Nevertheless, as discussed above, many existing diffusion pipelines for S1/S2 reconstruction still assume paired inputs or depend on preprocessing alignment rather than performing timestamp-conditioned alignment internally.

These gaps motivate our approach, which integrates (i) internal timestamp-conditioned alignment for irregular and asynchronous S1/S2 fusion, (ii) spatially aware spatiotemporal modeling (beyond per-pixel trajectories), and (iii) anytime querying to synthesize cloud-free S2 at both observed and user-specified timestamps within a monitoring window. An overview of AGFlow pipeline is shown in Figure~\ref{fig:teaser}.

\section{Method}
\label{sec:method}
AGFlow learns a continuous-time vector field that reconstructs missing values in a satellite image time series while clamping all observed pixels.
This single formulation covers (i) \textbf{cloud/sensor-gap removal} (pixel-level masks) and (ii) \textbf{anytime generation} (full-frame masks), by changing only the missingness mask.

\subsection{Problem setup}
\label{sec:setup}

Let $\mathbf{x}\in\mathbb{R}^{T\times C\times H\times W}$ denote an optical sequence (S2) of length $T$, with $C$ bands and spatial size $H\times W$.
We assume an underlying clean sequence $\mathbf{y}\in\mathbb{R}^{T\times C\times H\times W}$.
A binary mask $\mathbf{m}\in\{0,1\}^{T\times 1\times H\times W}$ indicates missingness, where
$m_{i,1,u,v}=1$ means pixel $(u,v)$ at frame $i$ is unknown and must be generated, and $m_{i,1,u,v}=0$ means it is observed.
The mask is broadcast across channels in all elementwise products.

The observed tensor $\mathbf{x}$ stores the measured values on known pixels and a constant fill value $v_{\text{fill}}$ on unknown pixels:
\begin{equation}
\mathbf{x} = (1-\mathbf{m})\odot \mathbf{y} \;+\; \mathbf{m}\odot v_{\text{fill}}\mathbf{1},
\end{equation}
where $\odot$ is elementwise multiplication and $\mathbf{1}$ is an all-ones tensor of appropriate shape.

\paragraph{SAR conditioning.}
We condition on a co-registered SAR sequence (S1)
$\mathbf{s}\in\mathbb{R}^{T_s\times C_s\times H\times W}$ with length $T_s$ and $C_s$ channels.
We also use acquisition dates (in days since a reference):
$\{d_i\}_{i=1}^T$ for optical and $\{d^{(s)}_j\}_{j=1}^{T_s}$ for SAR.
Our goal is to predict $\mathbf{y}$ on masked pixels ($\mathbf{m}=1$) while preserving all observed pixels ($\mathbf{m}=0$).

\subsection{Unifying cloud removal and anytime generation}
\label{sec:anytime}

AGFlow treats both tasks as masked reconstruction.

\textbf{Cloud/sensor-gap removal.}
In the standard setting, $\mathbf{m}$ encodes pixel-level missing regions (clouds, shadows, sensor gaps), so the model inpaints only those pixels.

\textbf{Anytime generation via full-frame masking.}
To enable synthesis of an arbitrary time step, during training we sample a \emph{query} index $q\in\{1,\dots,T\}$ and set the entire query frame to unknown:
\begin{equation}
\mathbf{m}_{q,:,:,:} \leftarrow \mathbf{1}_{1\times H\times W},
\qquad
\mathbf{x}_{q,:,:,:} \leftarrow v_{\text{fill}}\,\mathbf{1}_{C\times H\times W}.
\label{eq:anytime}
\end{equation}
Thus, the model must reconstruct the full query frame from the remaining temporal context (and SAR), while still handling pixel-level missingness in other frames.
At test time, anytime generation is performed by setting the desired frame(s) to $\mathbf{m}=1$ (even if the frame is fully missing).
\subsection{Masked Flow Matching}
\label{sec:fm}

AGFlow learns a vector field that transports samples from noise to data, while restricting learning and updates to masked pixels.

For each training sample, we draw Gaussian noise
$\boldsymbol{\epsilon}\sim\mathcal{N}(\mathbf{0},\mathbf{I})$
with the same shape as $\mathbf{y}$, and sample a continuous flow time $\tau\sim\mathcal{U}(0,1)$.
We define a linear path between data and noise:
\begin{equation}
\mathbf{y}_{\tau} = (1-\tau)\mathbf{y} + \tau\boldsymbol{\epsilon}.
\label{eq:fm_path}
\end{equation}
Differentiating gives the target velocity field (constant along the path):
\begin{equation}
\mathbf{v}^\star = \frac{d\mathbf{y}_{\tau}}{d\tau} = \boldsymbol{\epsilon} - \mathbf{y}.
\label{eq:fm_vstar}
\end{equation}

To ensure observed pixels are clamped in the network input, we replace known regions by $\mathbf{x}$:
\begin{equation}
\mathbf{z}_{\tau} = \mathbf{m}\odot \mathbf{y}_{\tau} + (1-\mathbf{m})\odot \mathbf{x}.
\label{eq:fm_inpaint}
\end{equation}

Let $\mathbf{v}_\theta(\cdot)$ denote the learned vector field.
We minimize mean-squared error \emph{only on masked pixels}:
\begin{equation}
\mathcal{L}(\theta)=
\mathbb{E}_{\mathbf{y},\,\boldsymbol{\epsilon},\,\tau}\left[
\left\|
\mathbf{m}\odot\left(
\mathbf{v}_\theta(\mathbf{z}_{\tau}, \tau, \mathrm{cond}) - (\boldsymbol{\epsilon}-\mathbf{y})
\right)
\right\|_F^2
\right],
\label{eq:fm_loss}
\end{equation}
where $\|\cdot\|_F$ is the Frobenius norm and $\mathrm{cond}$ includes dates and optional SAR.

\paragraph{Sampling with fixed observed pixels.}
At inference, we start from $\tilde{\mathbf{y}}_{1}=\boldsymbol{\epsilon}$ and solve the ordinary differential equation (ODE) backward from $\tau=1$ to $\tau=0$:
\begin{align}
\tilde{\mathbf{z}}_{\tau} &= \mathbf{m}\odot \tilde{\mathbf{y}}_{\tau} + (1-\mathbf{m})\odot \mathbf{x}, \\
\frac{d\tilde{\mathbf{y}}_{\tau}}{d\tau} &= \mathbf{m}\odot \mathbf{v}_\theta(\tilde{\mathbf{z}}_{\tau}, \tau, \mathrm{cond}).
\label{eq:masked_sampling_ode}
\end{align}
Because the RHS is multiplied by $\mathbf{m}$, all observed pixels ($\mathbf{m}=0$) have zero velocity and remain fixed for any numerical solver.

We parameterize $\mathbf{v}_\theta$ with Sequential Denoising Transformer (SDT), a Transformer over spatio-temporal patch tokens.
With patch size $p\times p$, each frame has $N=\frac{HW}{p^2}$ patches.
A linear patch embedding maps each patch to a token of width $M$.
For each frame $i$, we obtain
\begin{equation}
\mathbf{H}^{(0)}_i = \mathrm{PatchEmbed}(\mathbf{z}_{\tau}[i]) \in \mathbb{R}^{N\times M}.
\end{equation}
We add a fixed 2D sine/cosine spatial embedding $\mathbf{P}\in\mathbb{R}^{N\times M}$:
\begin{equation}
\mathbf{H}^{(0)}_i \leftarrow \mathbf{H}^{(0)}_i + \mathbf{P}.
\end{equation}
Stacking over time yields $\mathbf{H}^{(0)}\in\mathbb{R}^{T\times N\times M}$.

We embed the continuous flow time $\tau$ via a sinusoidal embedding followed by an MLP:
\begin{equation}
\mathbf{c}=\psi(\tau)\in\mathbb{R}^{M}.
\end{equation}
We also encode acquisition dates. Let $\phi(\cdot)$ be a date encoder producing vectors in $\mathbb{R}^{M}$.
We compute absolute and (optionally) query-relative date embeddings:
\begin{equation}
\mathbf{e}^{\text{abs}}_i=\phi(d_i), 
\qquad
\mathbf{e}^{\Delta}_i=\phi(d_i-d_q),
\end{equation}
and combine them with learned scalars
\begin{equation}
\mathbf{e}_i=\lambda_{\text{abs}}\mathbf{e}^{\text{abs}}_i+\lambda_{\Delta}\mathbf{e}^{\Delta}_i \in\mathbb{R}^{M}.
\label{eq:date_emb}
\end{equation}
The resulting $\mathbf{e}_i$ is added to all patch tokens of frame $i$.

Each Transformer block is conditioned on $\mathbf{c}$ using adaptive LayerNorm (AdaLN) \cite{peebles2023scalable} with gating:
\begin{align}
\mathrm{AdaLN}(\mathbf{X};\mathbf{c}) &= (1+\gamma(\mathbf{c}))\odot \mathrm{LN}(\mathbf{X})+\beta(\mathbf{c}),\\
\mathbf{X} &\leftarrow \mathbf{X} + g(\mathbf{c})\odot f\!\big(\mathrm{AdaLN}(\mathbf{X};\mathbf{c})\big),
\label{eq:adln}
\end{align}
where $\gamma(\cdot)$, $\beta(\cdot)$, and $g(\cdot)$ are learned linear projections of $\mathbf{c}$ and $f(\cdot)$ denotes an attention or MLP sublayer.

\subsection{Time-Aligned cross-attention for SAR fusion}
\label{sec:sar}
We tokenize $\mathbf{s}$ analogously. For each SAR frame $j$:
\begin{equation}
\mathbf{G}^{(0)}_j = \mathrm{PatchEmbed}(\mathbf{s}[j]) + \mathbf{P} + \mathbf{e}^{(s)}_j
\in \mathbb{R}^{N \times M},
\end{equation}
where $\mathbf{e}^{(s)}_j=\phi(d^{(s)}_j)$ is broadcast to all SAR patch tokens.
Stacking gives $\mathbf{G}^{(0)}\in\mathbb{R}^{T_s\times N\times M}$.

We fuse SAR features into SDT using \textbf{Time-Aligned Cross-Attention} consisting of:

\paragraph{(i) Spatial cross-attention (frame-wise).}
For each optical frame $i$, optical patch tokens attend to SAR patch tokens (same spatial grid), which helps match local structures across sensors.

\paragraph{(ii) Temporal cross-attention (patch-wise).}
For a fixed patch index $n\in\{1,\dots,N\}$, we reshape sequences across time:
\begin{equation}
\tilde{\mathbf{H}}_n \in\mathbb{R}^{T\times M}, \qquad
\tilde{\mathbf{G}}_n \in\mathbb{R}^{T_s\times M}.
\end{equation}
Multi-head cross-attention is
\begin{equation}
\mathrm{Attn}(\mathbf{Q},\mathbf{K},\mathbf{V})
=
\mathrm{softmax}\!\left(
\frac{\mathbf{Q}\mathbf{K}^\top}{\sqrt{d_h}} + \mathbf{B}^{\text{S2}\leftarrow\text{S1}}_{\Delta d}
\right)\mathbf{V},
\label{eq:temporal_xattn}
\end{equation}
where $\mathbf{Q}=\tilde{\mathbf{H}}_n\mathbf{W}_Q$, $\mathbf{K}=\tilde{\mathbf{G}}_n\mathbf{W}_K$, $\mathbf{V}=\tilde{\mathbf{G}}_n\mathbf{W}_V$, $d_h$ is the head dimension, and $\mathbf{B}^{\text{S2}\leftarrow\text{S1}}_{\Delta d}$ is a learned bias based on cross-sensor day differences.
This explicitly encourages matching the most relevant SAR times for each optical time.

\subsection{Real-date temporal encoding}
\label{sec:timebias}
AGFlow injects real acquisition times only into attention layers that mix across time:
(i) temporal self-attention and (ii) temporal cross-attention (Eq.~\eqref{eq:temporal_xattn}).

\paragraph{Relative time bias (within optical).}
For temporal attention within the optical sequence, we add a learned bias that depends on day differences:
\begin{equation}
\mathrm{Attn}(\mathbf{Q},\mathbf{K},\mathbf{V})
=
\mathrm{softmax}\!\left(
\frac{\mathbf{Q}\mathbf{K}^\top}{\sqrt{d_h}} + \mathbf{B}_{\Delta d}
\right)\mathbf{V},
\label{eq:attn_bias}
\end{equation}
with:
\begin{equation}
\mathbf{B}_{\Delta d}[h,i,k] = b_h\!\left(\rho(\Delta d_{ik})\right),
\qquad
\Delta d_{ik} = d_i - d_k,
\label{eq:bias_bucket}
\end{equation}
where $\rho(\cdot)$ buckets day differences and $b_h(\cdot)$ is a learned embedding for head $h$.

\textbf{Cross-sensor bias (optical $\leftarrow$ SAR).}
For temporal cross-attention from SAR to optical, we use
\begin{equation}
\Delta d_{i,j}=d_i-d^{(s)}_{j},
\qquad
\mathbf{B}^{\text{S2}\leftarrow\text{S1}}_{\Delta d}[h,i,j]
=
b_h\!\left(\rho(\Delta d_{i,j})\right).
\label{eq:cross_sensor_bias}
\end{equation}

\textbf{Rotary temporal encoding (RoPE).}
In addition, we apply RoPE \cite{su2024roformer} to temporal attention by rotating $\mathbf{Q}$ and $\mathbf{K}$ using time positions derived from real dates (day values).
This gives smooth, continuous phase shifts for irregular sampling, while the learned biases handle coarse alignment and sensor-specific timing.

\section{ Experiments \& results}
\label{sec:exp}
\subsection{Dataset, Protocol, and Preprocessing}
\paragraph{Dataset and sensors.}
We follow the RESTORE-DiT benchmark and use S2  optical time series as the reconstruction target, with S1 SAR time series as an all-weather condition signal. S2 uses atmospherically corrected products with 10 spectral bands (excluding atmospheric bands). S1 uses Ground Range Detected (GRD) products processed to backscatter (dB) and orthorectified to 10\,m resolution. We represent S1 with three channels (VV, VH, and VV/VH) and use ascending-orbit acquisitions for consistency.
\begin{figure}
    \centering
    \includegraphics[width=\linewidth]{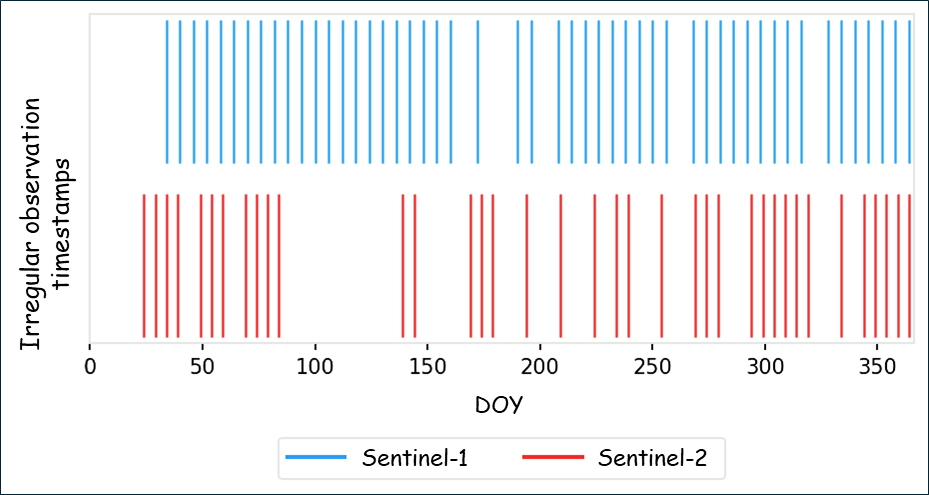}
    \caption{ The irregular and asynchronous distribution of timestamps across both sensors reflects the combined effect of satellite revisit cycles, orbit geometry, and cloud-induced data gaps in the optical record.}
    \label{fig:irregular}
\end{figure}

\textbf{Regions and splits.}
Training and main evaluation are on the France site from PASTIS-R \cite{garnot2022multi}, collected from four tiles (30UXV, 32ULU, 31TFM, 31TFJ). After cropping and filtering, the dataset contains 2433 non-overlapping patches of size $128\times128$ pixels ($\approx 1.28\times1.28$\,km). The temporal span covers more than one year (Sep 2018--Nov 2019), with 38--61 S2 acquisitions per series and $\sim$70 S1 acquisitions per series. We use the official 5-fold split: four folds for training and one fold for testing. 

During training, we simulate missingness by sampling masks from the real mask pool and overlaying them on the cloud-free sequences, so the missing-data distribution matches real cloud patterns. This produces (i) a masked S2 observation sequence, (ii) a mask sequence indicating missing pixels, and (iii) the corresponding cloud-free S2 target sequence.


\textbf{Normalization and windowing.}
We clip S2 values to a fixed reflectance range, standardize S1 with dataset statistics, and rescale both modalities to a common normalized range for training. We use fixed-length temporal windows (length 15); longer sequences are processed with a sliding-window strategy.

\textbf{Evaluation protocol and metrics.}
We report standard reconstruction metrics: mean absolute error (MAE), root mean squared error (RMSE), spectral angle mapper (SAM), peak signal-to-noise ratio (PSNR), and structural similarity index (SSIM). For pixel-wise metrics, we evaluate only on missing/occluded pixels; SSIM is computed on frames that contain missing regions. We evaluate (i) cloud/sensor-gap removal where masks are pixel-level, and (ii) missing-frame reconstruction where a full S2 frame is removed and reconstructed from the remaining temporal context (and SAR). Anytime querying is evaluated by fully masking the queried frame(s) at test time. Moreover, all baselines and AGFlow are trained from scratch and evaluated on the same official split (train folds vs the held-out test fold).

\textbf{Irregular temporal sampling.}
The dataset comprises multi-source satellite time series with irregular temporal sampling (Figure~\ref{fig:irregular}).
Observations from S1 and S2 are acquired at different and non-uniform intervals, resulting in asynchronous timestamp lists across sensors.
The Day of Year (DOY) axis is anchored to September 1, 2018 as the reference date, spanning a full seasonal cycle.

\subsection{Implementation details.}
We use an SDT with depth 4, hidden size 256, patch size $4\times4$, 4 attention heads, and MLP ratio 4.0, operating on $128\times128$ crops with a fixed temporal window $T=15$. Inputs are S2 (10 bands) and S1 (3 channels). S2 values are clipped to $[0,8000]$; S1 is standardized using dataset statistics, clipped to $[-2,2]$. Cloud masks are sampled from the provided real mask pool; masked pixels are filled with value $1$ (in normalized space). For anytime training, the query frame is fully masked and the loss is the MSE on masked pixels using flow matching with ${\tau}\sim\mathcal{U}(0,1)$, linear interpolation $y_{\tau}=(1-{\tau})y+{\tau}\epsilon$, and target velocity $\epsilon-y$. We train with Adam (lr $2\times 10^{-4}$, batch size 16) for 3000 epochs and use a MultiStepLR schedule.

\begin{table}
\centering
\small
\setlength{\tabcolsep}{4pt}
\begin{tabular}{l c c c c c}
\toprule
Model & $\downarrow$MAE & $\downarrow$RMSE & $\downarrow$SAM & $\uparrow$PSNR & $\uparrow$SSIM \\
\midrule
RESTORE-DiT & 0.0214 & 0.0322 & 2.95 & 32.17 & 0.914 \\
\textbf{AGFlow (full)} & \textbf{0.0179} & \textbf{0.0261} & \textbf{2.78} & \textbf{32.87} & \textbf{0.942} \\
\bottomrule
\end{tabular}
\vspace{2pt}
\caption{Missing-frame (gap filling) results. One S2 frame is fully removed at a random time index and reconstructed from the remaining temporal context. }
\label{tab:missing}
\end{table}

\subsection{Missing frame / gap filling}
For each test sequence, we randomly choose one time index $t$ and remove the corresponding S2 frame (all bands, all pixels).
The model then generates the missing frame $\hat{\mathbf{x}}_t$ conditioned on the remaining observed S2 frames and all S1 frames.
We compare $\hat{\mathbf{x}}_t$ with the held-out ground truth $\mathbf{x}_t$ (Table~\ref{tab:missing}).

Table~\ref{tab:missing} shows that AGFlow outperforms the diffusion baseline RESTORE-DiT on all metrics.
AGFlow reduces MAE from 0.0214 to 0.0179 ($-16.4\%$) and RMSE from 0.0322 to 0.0261 ($-18.9\%$),
indicating lower overall reconstruction error when the target frame is missing.
It also improves spectral consistency (SAM: 2.9514 $\rightarrow$ 2.7761) and image quality (PSNR: 32.1755 $\rightarrow$ 32.8671; SSIM: 0.9139 $\rightarrow$ 0.9420).
These results support that AGFlow's temporally-aligned conditioning better leverages surrounding observations in the fully-missing-frame setting.

\begin{figure*}[t]
    \centering
    \includegraphics[width=\linewidth]{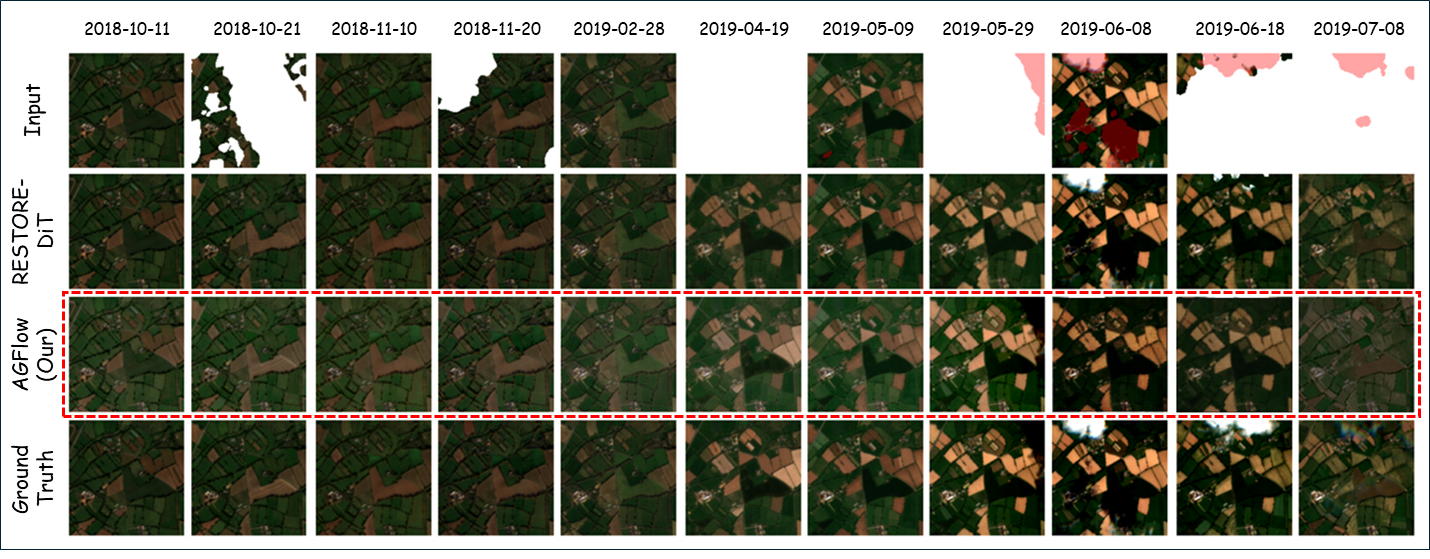}
    \vspace{-2mm}
    \caption{Missing-frame reconstruction and cloud removal.
The top row shows degraded S2 inputs with masks indicating invalid pixels: white regions indicate total occlusion (such as thick clouds or missing data), while pink and dark red regions denote secondary artifacts like cloud shadows or thin clouds. AGFlow demonstrates superior performance in heavily obscured frames. Notably, on dates 2019-06-08 and 2019-06-18, RESTORE-DiT struggles with spatial distortion and leaves residual cloudy artifacts. In contrast, AGFlow produces reconstructions with fewer artifacts and sharper structures, and it reduces cloud residuals when masked frames contain clouds.}
    \label{fig:france_cloud_vis}
    \vspace{-2mm}
\end{figure*}

\begin{table}
\centering
\small
\setlength{\tabcolsep}{4pt}
\begin{tabular}{l l c r r r r r}
\toprule
Methods   & $\downarrow$MAE & $\downarrow$RMSE & $\downarrow$SAM & $\uparrow$PSNR & $\uparrow$SSIM \\
\midrule
Linear &  0.0257 & 0.0401 & 4.35 & 28.40 & 0.929 \\
U-TILISE &  0.0202 & 0.0314 & 3.76 & 30.38 & 0.936 \\
U-TILISE-SAR &  0.0193 & 0.0298 & 3.66 & 30.77 & 0.937 \\
RESTORE-DiT & \underline{0.0140} & \underline{0.0224} & \underline{2.64} & \underline{33.32} & \underline{0.959} \\
\textbf{AGFlow (ours)} & \textbf{0.0133} & \textbf{0.0217} & \textbf{2.45} & \textbf{33.65} & \textbf{0.964} \\
\bottomrule
\end{tabular}
\vspace{2pt}
\caption{
Quantitative evaluation on the France test set for cloud removal.
Metrics are computed over all ten S2 bands and only on cloud-corrupted pixels.
Best and second-best results are highlighted in \textbf{bold} and \underline{underline}, respectively.
Linear and U-TILISE use optical inputs only; U-TILISE-SAR, RESTORE-DiT, and AGFlow use optical+SAR.
}
\label{tab:france_quant}
\end{table}

\subsection{Cloud removal}

Table~\ref{tab:france_quant} reports quantitative results on the France test set.

Methods that use temporal inputs outperform the single-image baseline (Linear), confirming that temporal context is important for cloud removal.
Among learning-based baselines, U-TILISE improves over Linear, and adding SAR (U-TILISE-SAR) yields a further but modest gain, suggesting SAR helps but simple concatenation does not fully exploit asynchronous SAR timing.

Diffusion-based restoration is strongest overall in this benchmark. RESTORE-DiT is the best baseline, and \textbf{AGFlow achieves the top performance}.
Compared to RESTORE-DiT, AGFlow improves MAE from 0.0140 to 0.0133 ($-5.0\%$), RMSE from 0.0224 to 0.0217 ($-3.1\%$), and SAM from 2.64 to 2.45 ($-7.2\%$).
It also slightly improves perceptual quality (PSNR: 33.32 $\rightarrow$ 33.65; SSIM: 0.959 $\rightarrow$ 0.964).
Overall, the gains are consistent and strongest on spectral fidelity (SAM) and SSIM.

Figure~\ref{fig:france_cloud_vis} provides a qualitative comparison against RESTORE-DiT and the ground truth.
In the shown example(s), AGFlow produces reconstructions that are closer to the ground truth with fewer visible artifacts in cloud-covered regions. For example, for images captured on 06-08-2019 and 06-18-2019, AGFlow is able to remove clouds based on inferences from the time sequence.

\subsection{Anytime generation}
\begin{figure*}
    \centering
    \includegraphics[width=1\linewidth]{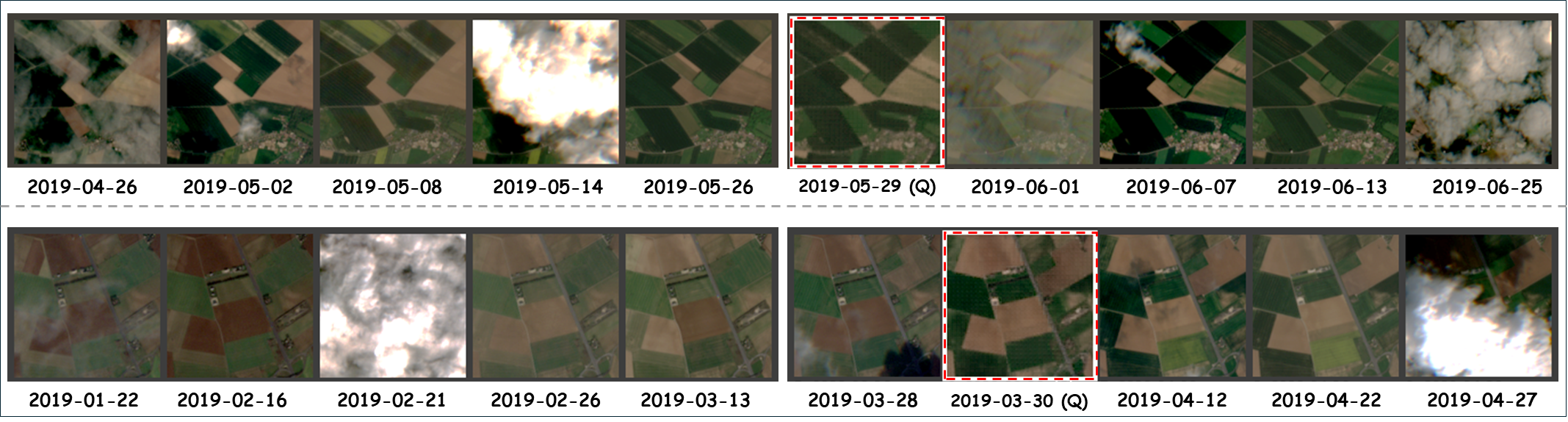}
    \caption{Anytime querying examples. For each sequence, we query AGFlow at dates marked (Q) that are not observed by S2. We show surrounding observed frames for context and the generated cloud-free output at the query date.}
    \label{fig:anytime}
\end{figure*}

\begin{figure*}
    \centering
    \includegraphics[width=0.85\linewidth]{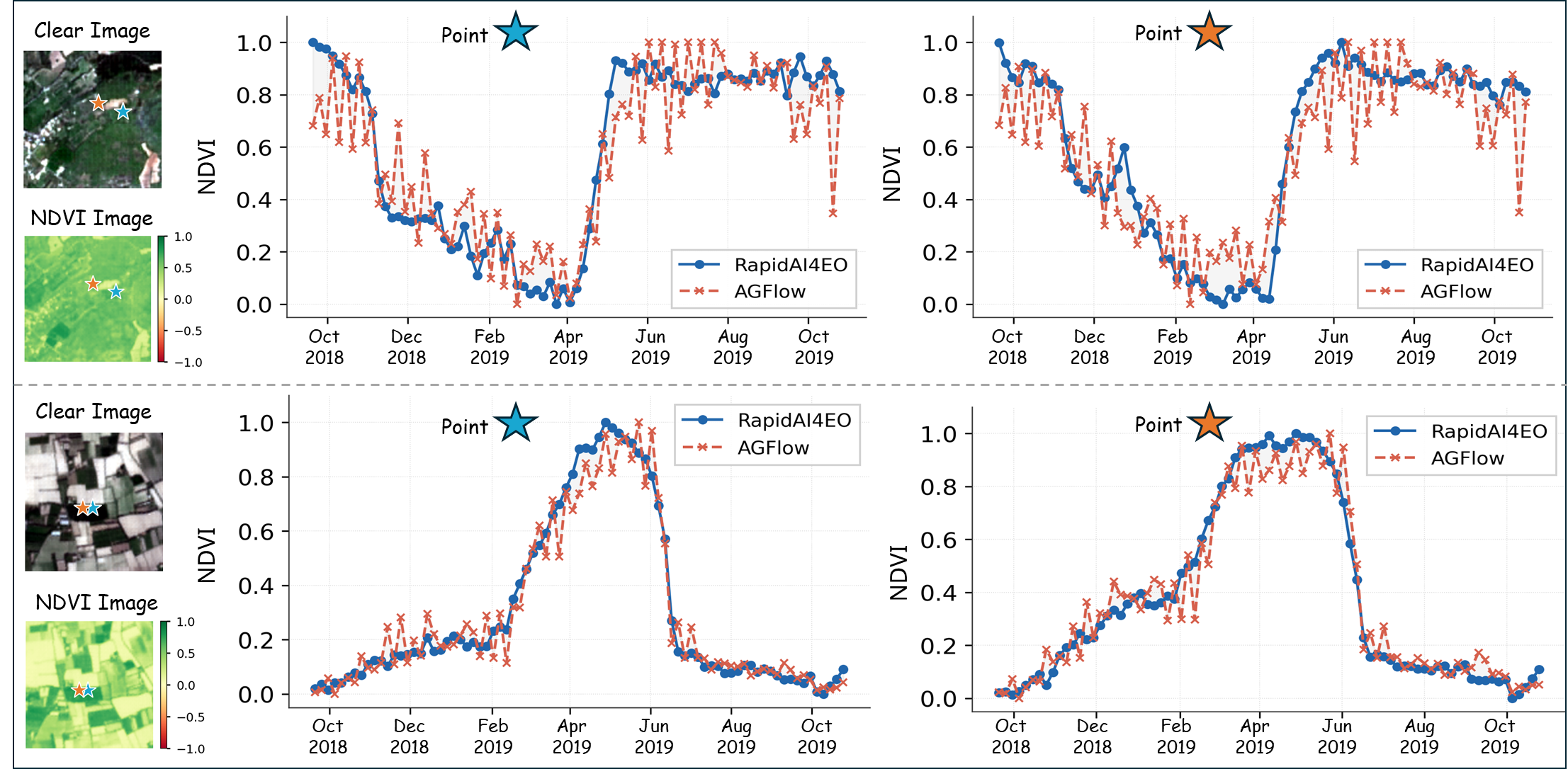}
    \caption{NDVI-based anytime evaluation against an auxiliary cloud-free reference (RapidAI4EO). We compute NDVI from AGFlow outputs at queried dates and compare it to NDVI from the auxiliary reference using regional summary statistics over time. The goal is to check whether generated seasonal dynamics are plausible and consistent at the region level, despite sensor and timestamp mismatch.}
    \label{fig:ndvi}
\end{figure*}

Figure~\ref{fig:anytime} shows qualitative examples where AGFlow generates cloud-free S2 outputs at unobserved query dates. In anytime generation, the model is asked to generate plausible optical observations for time steps where no optical measurement is available. In our target setting, there is no aligned ground-truth S2 frame for these generated timestamps, so direct supervised evaluation is not possible. 

To still quantify the outputs, we follow the evaluation strategy used in AnytimeFormer \cite{tang2026anytimeformer} and leverage the RapidAI4EO corpus \cite{rapidai4eo}. RapidAI4EO is a large-scale European time-series dataset sampled at 500,000 non-overlapping locations (each covering a $\sim$600\,m $\times$ 600\,m footprint) and provides analysis-ready, regular-cadence products designed for land monitoring. In particular, RapidAI4EO pairs (i) S2 mosaics (monthly) with (ii) a high-cadence, cloud-free harmonized product (Planet Fusion) over the same locations, along with land-cover labels (CORINE/CLC) \cite{rapidai4eo}. In our experiments, we use the RapidAI4EO subset covering the same geographic region as our test site, and treat the cloud-free product as an auxiliary reference for evaluating temporal vegetation dynamics.

We compute Normalized Difference Vegetation Index (NDVI) from the generated S2 images and compare it against NDVI computed from the RapidAI4EO auxiliary reference. We use the standard NDVI definition:
\begin{equation}
\mathrm{NDVI} = \frac{\rho_{\mathrm{NIR}} - \rho_{\mathrm{Red}}}{\rho_{\mathrm{NIR}} + \rho_{\mathrm{Red}}},
\end{equation}
where $\rho_{\mathrm{NIR}}$ and $\rho_{\mathrm{Red}}$ are the near-infrared (NIR) and red reflectances.

Although RapidAI4EO provides observations over the same locations, the auxiliary reference is not perfectly aligned with our generated timestamps and differs in sensing/processing (cross-sensor harmonization, mosaicking, and product-specific pipelines). Because of this domain and timing gap, a strict pixel-wise comparison is not meaningful. Instead, we compare NDVI in a more robust way using regional summary statistics and temporal trend agreement (Figure~\ref{fig:ndvi}), reporting how well the generated NDVI distribution and regional dynamics match the auxiliary reference.

\begin{table*}
\centering
\small
\setlength{\tabcolsep}{4pt}
\begin{tabular}{l c c c c c}
\toprule
Model (variant) & $\downarrow$MAE & $\downarrow$RMSE & $\downarrow$SAM & $\uparrow$PSNR & $\uparrow$SSIM \\
\midrule
w/ spatial-only SAR fusion & 0.0190 & 0.0274 & 3.39 & 32.13 & 0.914 \\
w/o relative time bias (Eq.~\ref{eq:attn_bias})      & 0.0187 & 0.0268 & 3.33 & 32.33 & 0.915 \\
w/o query-relative timing ($\lambda_{\Delta}=0$)     & 0.0182 & 0.0264 & 2.79 & 32.78 & 0.941 \\
AGFlow (full)     & \textbf{0.0179} & \textbf{0.0261} & \textbf{2.77} & \textbf{32.86} & \textbf{0.942} \\
\bottomrule
\end{tabular}
\vspace{2pt}
\caption{Ablation results for missing-frame reconstruction. All variants share the same SDT backbone, training budget, and data.}
\label{tab:ablation_main}
\end{table*}

\subsection{Ablation study}
\label{sec:ablation}
We design ablations to isolate the main components of AGFlow beyond the SDT backbone.
All variants use the same SDT architecture, training schedule, and training data.

\textbf{Internal time alignment (relative time bias).}
We remove the learned relative time bias used in temporal attention (Eq.~\ref{eq:attn_bias}).
This tests whether explicit alignment cues inside attention are necessary when learning from irregular time gaps.

\textbf{Query-relative timing for anytime.}
We remove the query-relative date embedding by setting $\lambda_{\Delta}=0$ in Eq.~\ref{eq:date_emb}
(i.e., the model only receives absolute dates). This tests whether conditioning on the time difference to the query date
improves the anytime setting.

\textbf{SAR fusion mechanism.}
We replace the Time-Aligned temporal cross-attention with a weaker fusion strategy (spatial-only cross-attention).
This removes temporal matching during fusion and tests whether gains truly come from temporally-aware SAR--optical fusion.

\textbf{Results and discussion.}
Table~\ref{tab:ablation_main} summarizes performance on missing-frame reconstruction.
The full AGFlow model (last row) is best on all metrics.
All ablations reduce performance, so the gains are not explained by the SDT backbone or training budget.

The largest degradation comes from replacing Time-Aligned fusion with spatial-only fusion.
Compared to the full model, MAE increases from 0.0179 to 0.0190, PSNR drops from 32.8671 to 32.1383,
and SSIM drops from 0.9420 to 0.9147. The spectral error also increases strongly (SAM: 2.7761 $\rightarrow$ 3.3959, +0.6198).
This indicates that temporal matching during SAR fusion is important not only for pixel accuracy (MAE/RMSE, PSNR/SSIM),
but also for preserving spectral consistency (SAM).

Removing the learned relative time bias also causes a clear regression, especially in SAM and SSIM
(SAM: 2.7761 $\rightarrow$ 3.3358, +0.5597; SSIM: 0.9420 $\rightarrow$ 0.9154, -0.0266),
while the increase in MAE/RMSE is smaller than the fusion ablation.
This pattern suggests that internal time alignment mainly helps the model choose the correct temporal evidence,
which is critical for spectral fidelity.

Finally, removing query-relative timing yields a smaller but consistent drop across all metrics
(e.g., MAE: 0.0179 $\rightarrow$ 0.0182, +0.0003; PSNR: 32.8671 $\rightarrow$ 32.7849, -0.0822).
This shows that query-relative conditioning provides an extra benefit for anytime reconstruction, even when absolute dates are available.

\section{Conclusion and Limitations}

\textbf{Conclusion.}
We presented AGFlow for multimodal S1/S2 time-series reconstruction under irregular sampling and asynchronous acquisition. AGFlow uses a unified masked formulation that covers both pixel-level cloud/sensor-gap removal and full-frame masking for query-time reconstruction, while preserving all observed pixels. Across the RESTORE-DiT benchmark protocol, AGFlow improves missing-frame reconstruction over the diffusion baseline and produces more reliable reconstructions under persistent gaps, supporting flexible downstream use such as dense vegetation monitoring.

\textbf{Limitations.}
First, the approach assumes reliable spatial co-registration between S1 and S2; misregistration can degrade fusion quality. Second, to control compute and memory we rely on fixed-length temporal windows; very long-range dependencies outside the window may be only partially captured. Finally, the method is trained and tuned on Sentinel-1/2 processing conventions; adapting to other sensors, resolutions, or regions with different scattering/reflectance behavior may require additional retraining and calibration.
\section*{Acknowledgment}
This research is supported in part by Google.org’s Impact Challenge for Climate Innovation Program and the National Science Foundation under award 2120943. We thank Google Cloud Platform and the Research Computing (RC) at Arizona State University (ASU) for their support in providing computing resources to support this research. 

{
    \small
    \bibliographystyle{ieeenat_fullname}
    \bibliography{main}
}


\end{document}